%% file: main.tex
\def\year{2023}\relax
\documentclass[letterpaper]{article} 
\usepackage{aaai22}  
\usepackage{times}  
\usepackage{helvet}  
\usepackage{courier}  
\usepackage[hyphens]{url}  
\usepackage{graphicx} 
\usepackage{natbib}  
\usepackage{multirow}
\usepackage{graphicx}
\usepackage{caption} 
\DeclareCaptionStyle{ruled}{labelfont=normalfont,labelsep=colon,strut=off} 
\usepackage{algorithm}
\usepackage{algorithmic}

\usepackage[table,xcdraw]{xcolor}

\usepackage{newfloat}
\usepackage{listings}
\floatstyle{ruled}
\newfloat{listing}{tb}{lst}{}
\floatname{listing}{Listing}
\title{The Evolution of Substance Use Coverage in the Philadelphia Inquirer}

\author{
Layla Bouzoubaa, \textsuperscript{\rm 1}
Ramtin Ehsani, \textsuperscript{\rm 1}
Preetha Chatterjee, \textsuperscript{\rm 1}
Rezvaneh Rezapour \textsuperscript{\rm 1}
}
\affiliations{
\textsuperscript{\rm 1} College of Computing and Informatics, Drexel University, Philadelphia, USA\\
\{lb3338, re378, preethac, shadi.rezapour\}@drexel.edu
}

\begin{document}

\maketitle

\begin{abstract}
The media's representation of illicit substance use can lead to harmful stereotypes and stigmatization for individuals struggling with addiction, ultimately influencing public perception, policy, and public health outcomes. To explore how the discourse and coverage of illicit drug use changed over time, this study analyzes 157,476 articles published in the Philadelphia Inquirer over a decade. Specifically, the study focuses on articles that mentioned at least one commonly abused substance, resulting in a sample of 3,903 articles. Our analysis shows that cannabis and narcotics are the most frequently discussed classes of drugs. Hallucinogenic drugs are portrayed more positively than other categories, whereas narcotics are portrayed the most negatively. Our research aims to highlight the need for accurate and inclusive portrayals of substance use and addiction in the media.
\end{abstract}

\input{intro.tex}
\input{methodology}
\input{results}
\input{conclusion}

\bibliography{ICWSM_data/references,ICWSM_data/aaai22}

\end{document}

%% file: intro.tex
\section{Introduction}
The portrayal of illicit substance use in the media has long been a topic of concern, given the potential impact it can have on public perception and policy. Previous studies highlighted the harmful effects of stigmatizing coverage on individuals struggling with substance use disorder (SUD) and the potential for biased and inaccurate media coverage to perpetuate harmful stereotypes and stigma \cite{johnston2004monitoring, caburnay2003news, hughes_how_2011}. Moreover, media coverage often prioritizes criminal justice actions over critical health concerns associated with substance use \cite{hayden_griffin_deconstructing_2023}. This bias has been further observed in the case of designer drugs such as ``bath salts,'' which were portrayed as an ``epidemic'' with harmful effects while neglecting relevant clinical research and mental health issues associated with them \cite{swalve_framing_2016}. As a result, health legislation was inadequately informed, and underlying issues related to substance use are not addressed.

Given the potential impact of media coverage, it is essential to carefully examine how the media frames discussions about illicit drug use and how that framing changes over time. In this study, we examine news articles from the Philadelphia Inquirer to understand changes in media coverage of illicit substance use over time. Philadelphia was chosen because it has a high rate of overdose deaths (40 per 100K) and is home to a significant open-air drug market \cite{odd-philly}. Analyzing local media coverage of SUD can reveal insights on its portrayal, identify reporting gaps, and inform policymaking and public education. This type of study can also track the policy process and evolving relationships between actors, like advocacy groups or healthcare professionals, over time \cite{paalman1997or}.
More specifically, we analyze the portrayal of nine main drug classes, including \textit{stimulants, narcotics, cannabis, hallucinogens, depressants, designer drugs, drugs of concern, treatment}, and \textit{miscellaneous} over a period of ten years and across 157,476 published articles. The study then focuses on approximately 3,903 articles that mention at least one commonly abused substance and applies various natural language processing (NLP) techniques to extract drug-related themes over time. We aim to answer the following research questions:

\begin{itemize}
    \item[-] RQ1: How do the occurrences of various drug classes change over time, and in what contexts are different substances co-mentioned?\\
    We use a list of commonly abused drugs, map them to drug classes, extract articles with at least one mention of drugs, and use TF-IDF to find salient words in the subsets of articles discussing co-occurring drug classes. 
    \item[-] RQ2: How has media coverage of illicit substance use evolved over time?\\
    Dynamic topic modeling and sentiment analysis are employed to capture the shift of prevalent themes and tone with respect to drug classes, over time.
\end{itemize}

Our analysis indicates a shift in the discussion of drug-related topics, with cannabis being the most frequently discussed drug class, followed by narcotics. Articles on narcotics and treatment drugs frequently co-occur, focusing primarily on criminality, legislature, and overdose. News articles discussing cannabis, hallucinogens, and treatment drugs tend to be positive while those about narcotics, stimulants, and depressants are negative.
The insights gained from this work can inform better public health messaging, affect drug policy decisions made by lawmakers, and enable medical professionals to create more targeted patient education and awareness programs while promoting the principles of harm reduction.

%% file: methodology.tex
\section{Methodology}
\paragraph{Data Collection}
We utilize \textit{ProQuest} for data collection and accessing news articles. \textit{ProQuest} is an online platform consisting of thousands of databases that provide access to a diverse set of publications, including journals and newspapers.
In order to collect data, it was necessary to log in to the \textit{ProQuest} database using our individualized credentials, after which we were able to extract/scrape the data.
We collect a total of 157,476 news articles from the Philadelphia Inquirer, covering the period from January 1, 2013, to December 31, 2022.
It's important to note that our dataset may not represent the complete set of articles published by the Philadelphia Inquirer, as \textit{ProQuest}'s database may not be comprehensive.
Our data comprises the complete text of news articles, alongside several metadata attributes such as date, author, title, links, and subject keywords.\footnote{The meta-data of news articles, list of drugs, and code can be found on \url{https://github.com/social-nlp-lab/drugs_in_inquirer}}

\paragraph{Drug Name Extraction}
We utilize the categorization provided by the \textit{National Institute on Drug Abuse} (NIDA), which comprises 28 drug categories, each with its corresponding commercial names. To minimize irrelevant articles, we manually verified drug names and removed ambiguous terms with multiple meanings, e.g., the term ``pot'' could refer to ``flowering pot'' or ``cooking pot'' in addition to the drug ``marijuana''. 
We then assign each drug to one of the nine classes established by the Drug Enforcement Administration (DEA) \cite{Drugs}. These classes include: 

\begin{itemize}
    \item \textit{Cannabis (C):} marijuana is a mind-altering (psychoactive) drug, produced by the cannabis sativa plant.``Cannabis'' and ``marijuana'' are often used interchangeably, resulting in the appearance of cannabis as both the drug class and the drug name.
    \item \textit{Depressants (D):} known to induce sleep, relieve anxiety and muscle spasms, and prevent seizures, e.g., barbiturates, and sedative-hypnotic substances like GHB.
    \item \textit{Designer Drugs (DD):} produced illicitly with a slightly altered chemical structure to mimic the pharmacological effects of controlled substances, e.g., synthetic marijuana or synthetic cathinones. 
    \item \textit{Drugs of Concern (DC):} unregulated drugs that can be harmful if abused, e.g., kratom and xylazine.
    \item \textit{Hallucinogens (H):} derived from plants and fungi and renowned for their capacity to modify human perception and mood, e.g., LSD, mushrooms, and ecstasy.
    \item \textit{Narcotics (N):} refers to opium, opium derivatives, and their semi-synthetic substitutes. ``Opioid'' is a more current and precise term to describe these drugs, e.g., heroin, OxyContin, codeine, morphine, and fentanyl.
    \item \textit{Stimulants (S):} drugs accelerating body's functions, e.g., methamphetamine, cocaine, and amphetamines.
    \item \textit{Treatment (T):} substances aiding the treatment of opioid addiction, e.g., methadone, Suboxone, and naloxone.
    \item \textit{Miscellaneous (M):} substances that can be abused but don't belong to any classes, e.g., steroids.
\end{itemize}

\begin{table}[t]
\centering
\resizebox{0.9\columnwidth}{!}{%
\begin{tabular}{|c|cccccccc|c|}
\hline
           & \textbf{C}        & \textbf{D}  & \textbf{DD} & \textbf{DC} & \textbf{H} &\textbf{ N}      & \textbf{S}  & \textbf{T}     & \textbf{Total} \\ \hline
2013  & \textbf{152} & 23  & 3  & 2  & 38  & 142          & 108 & 16  & 484 \\ 
2014  & \textbf{150} & 10  & 0  & 1  & 53  & 117          & 64  & 21  & 416   \\
2015  & \textbf{132} & 6   & 0  & 0  & 35  & 117          & 49  & 30  & 369 \\ 
2016  & \textbf{152} & 12  & 0  & 0  & 44  & 132          & 63  & 43  & 446 \\ 
2017  & 203          & 12  & 0  & 0  & 37  & \textbf{223} & 53  & 78  & 606 \\ 
2018  & 226          & 14  & 8  & 6  & 43  & \textbf{228} & 73  & 98  & 696   \\ 
2019  & \textbf{234} & 10  & 4  & 2  & 55  & 148          & 90  & 48  & 591 \\ 
2020  & \textbf{119} & 7   & 0  & 0  & 18  & 69           & 48  & 31  & 292 \\ 
2021  & \textbf{124} & 7   & 0  & 0  & 40  & 86           & 41  & 41  & 339  \\
2022  & \textbf{145} & 5   & 1  & 0  & 31  & 99           & 38  & 43  & 362 \\ \hline
\textbf{Total} & 1637         & 106 & 16 & 11 & 394 & 1361         & 627 & 449 &     \\ \hline
\end{tabular}
}
\caption{Number of drug class occurrences per year, with bold values indicating the most frequent drug class per year. Drug classes include C (Cannabis), D (Depressants), DD (Designer Drugs), DC (Drugs of Concern), H (Hallucinogens), N (Narcotics), S (Stimulants), and T (Treatment).}
\label{tab:table1}
\end{table}

\begin{figure}[t]
    \centerline{\includegraphics[scale=0.265]{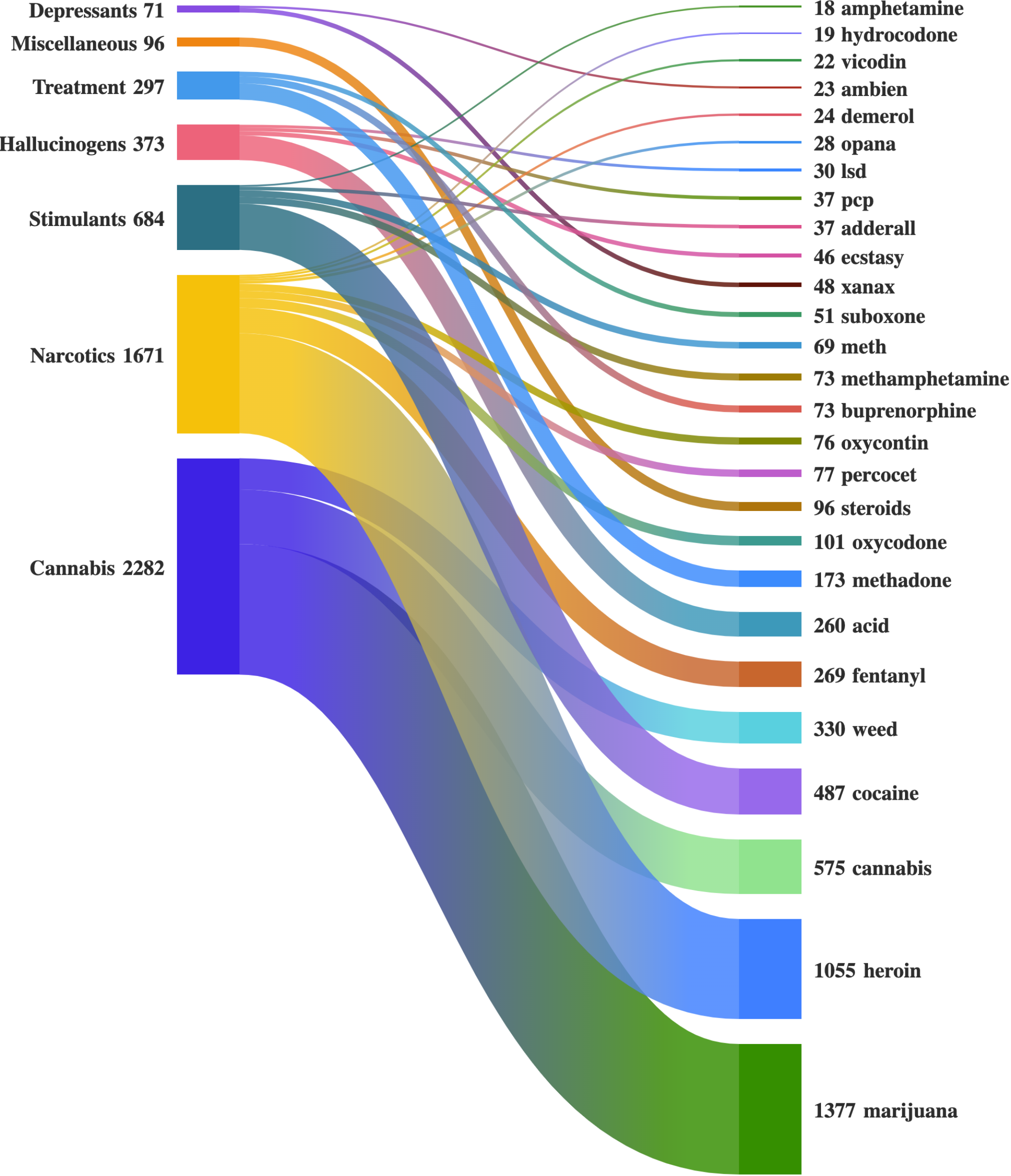}}
    \caption{Number of substances per drug class in our dataset.}
    \label{fig:snakey}
\end{figure}

\begin{figure*}[t]
    \centerline{\includegraphics[scale=0.2]{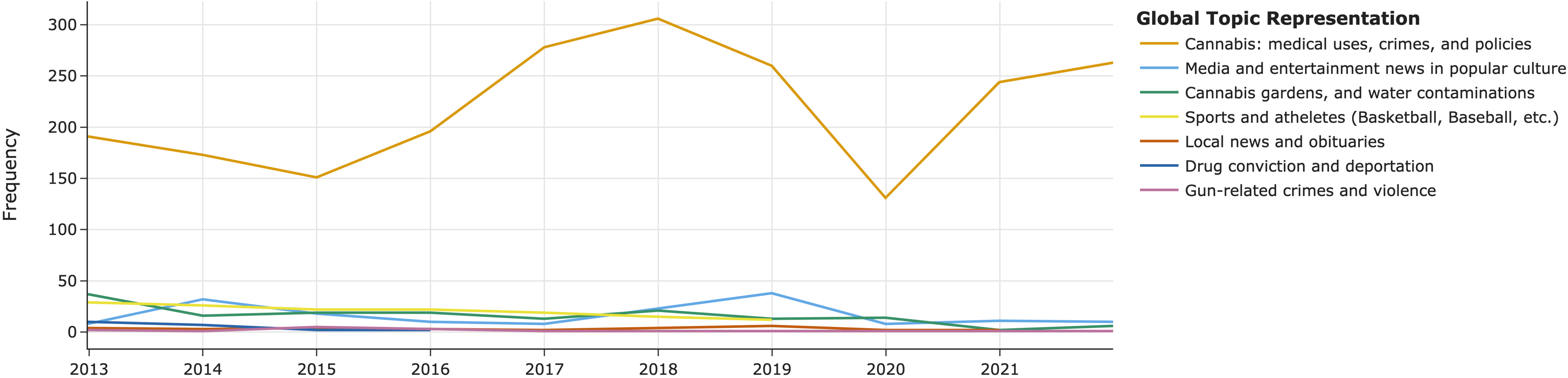}}
    \caption{News topic evolution over time, with each line representing a cluster of related topics.}
    \label{fig:topics_over_time}
\end{figure*}

Using our mapped list of drug names, we identify news articles that mention any of these drugs in either their title or body. Subsequently, we group these articles into clusters based on the respective drug classes. 
If an article mentions drugs from multiple classes, we assign it to all relevant drug classes. The number of articles per drug class and year is shown in Table \ref{tab:table1}. Out of a total of 157,476 articles analyzed, 3,903 referenced at least one drug class within their text.

In addition to assessing the frequency of drugs, we also examine co-occurrences of drugs in the same articles each year.
To better understand these co-occurrences and the surrounding contexts, we extract the most significant words based on their TF-IDF scores from a subset of articles where drugs most frequently co-occurred.

\paragraph{Topic Modeling}
To investigate the evolution of drug-related articles, we employ dynamic topic modeling \cite{blei2006dynamic} to understand the distribution of topics over time. We use BERTopic as it can preserve the semantic structure of texts \cite{grootendorst2022bertopic}. 
Since BERTopic operates on a bag-of-words representation of the text \cite{grootendorst2022bertopic}, we preprocess the text by converting it to lowercase, removing URLs, extending contractions, and eliminating stopwords. This minimizes noise and dimensionality, allowing our models to recognize important patterns and topics.
It is important to note that BERT has a token limit of 512, while the median length of news articles in our data is 749 tokens with a standard deviation of 573. News articles' opening paragraphs contain the most important information, therefore, we assume the first 512 tokens of news articles in our data represent the key points. 

\paragraph{Tone Detection}
VADER \cite{hutto_vader_2014}, a validated rule-based model for sentiment analysis, is used to analyze shifts in the tone of news articles that center on substance use. We calculate the average sentiment scores of articles in six-month periods using VADER's compound score, allowing us to monitor changes in sentiment over time.

%% file: results.tex
\section{Results}

\paragraph{How do the occurrences of various drug classes change over time, and in what contexts are different substances co-mentioned?}
The results show a significant shift in the discussion of illicit substance use over the past decade.
As shown in Table \ref{tab:table1}, cannabis is the most frequently discussed drug class, followed by narcotics. The annual count of news articles indicates that 2018 saw the highest frequency of drug-related mentions in news coverage. In 2018, several significant events occurred, including a former player for the Philadelphia Eagles admitting to recreational cannabis use during his career \cite{nfl2019}, Philadelphia Mayor Jim Kenney advocating for the legalization of recreational cannabis use \cite{mayor2018}, and the City of Philadelphia hosting its first-ever Cannabis Opportunity Conference \cite{brown_over_2018}. The frequency of cannabis in the news doesn't necessarily imply that it is used more than other substances, as it could be affected by the prevailing social and political environment. Our findings suggest that the majority of articles mentioning cannabis are centered around the legality of its use. 
Narcotics (i.e., opioids) are the second most discussed drug class, and also the most prevalent in 2017 and 2018, indicating the severity of discussions and possibly the opioid crisis during those years. Philadelphia experienced some of the highest rates of unintentional overdose rates, and the synthetic opiate fentanyl became increasingly fatal \cite{unintentional}.
As shown in Table \ref{tab:table1}, \textit{designer drugs} and \textit{drugs of concern} are mentioned in only 16 and 11 articles, respectively. Due to the sparse data on these topics in our corpus, we exclude them from further analysis. 

Further analysis of drug counts per class shows that cannabis is primarily associated with marijuana (Figure \ref{fig:snakey}). 
Despite the controversy surrounding the use of the term ``marijuana'' \cite{halperin_marijuana_2018}, there has been a consistent and steady usage of this word in news articles over the years.
In contrast, depressant drugs like Xanax and Ambien are more frequently mentioned together, likely due to their widespread use for anxiety and sleep disorders and their potential for abuse and addiction. Similarly, cocaine and acid are the most frequently mentioned drugs in the stimulant and hallucinogen drug classes, respectively, indicating their enduring popularity and widespread use.
It is worth noting that our temporal analysis of drug mentions uncovers patterns in drug discussion over time, which can be influenced by a variety of factors, such as changes in drug policies and cultural shifts in attitudes towards various elements of drug use, like policy, treatment, and research. 

\begin{table}[t]
\resizebox{\columnwidth}{!}{%
\begin{tabular}{|c|c|c|c|c|c|c|c|}
\hline
                      & Drug 1       & Drug 2    & N  &                       & Drug 1        & Drug 2        & N   \\ \cline{1-8} 
\multirow{3}{*}{2013} & heroin       & methadone & 52 & \multirow{3}{*}{2018} & fentanyl      & cocaine       & 270 \\
                      & heroin       & cocaine   & 45 &                       & heroin        & cocaine       & 194 \\
                      & barbiturates & cannabis  & 16 &                       & heroin        & marijuana     & 127 \\ \hline
\multirow{3}{*}{2014} & heroin       & naloxone  & 42 & \multirow{3}{*}{2019} & heroin        & marijuana     & 115 \\
                      & marijuana    & cocaine   & 36 &                       & heroin        & naloxone      & 77  \\
                      & heroin       & cocaine   & 34 &                       & heroin        & cocaine       & 71  \\ \hline
\multirow{3}{*}{2015} & heroin       & marijuana & 40 & \multirow{3}{*}{2020} & heroin        & methadone     & 198 \\
                      & heroin       & methadone & 19 &                       & heroin        & buprenorphine & 81  \\
                      & heroin       & meth      & 11 &                       & fentanyl      & cocaine       & 53  \\ \hline
\multirow{3}{*}{2016} & heroin & marijuana & 51  & \multirow{3}{*}{2021} & buprenorphine & cannabis & 108 \\ 
                      & heroin       & cocaine   & 45 &                       & heroin        & cannabis      & 78  \\
                      & heroin       & naloxone  & 44 &                       & heroin        & marijuana     & 41  \\ \hline
\multirow{3}{*}{2017} & heroin & methadone & 157 & \multirow{3}{*}{2022} & buprenorphine & cannabis & 351 \\
                      & heroin       & cocaine   & 90 &                       & buprenorphine & marijuana     & 153 \\ 
                      & heroin       & naloxone  & 80 &                       & fentanyl      & cocaine       & 114 \\ \hline
\end{tabular}%
}
\caption{The most frequent co-occurred drugs per year.}
\label{tab:co-occur}
\end{table}

The co-occurrence analysis (Table \ref{tab:co-occur}) revealed that certain substances are frequently discussed together, indicating a connection between their use, among other possible associations. 
Heroin was once the most discussed substance in Philadelphia until it was surpassed by fentanyl in 2018. This is in line with reports of adulteration and an increase in unintentional overdose deaths containing fentanyl in Philadelphia \cite{dea-engage-philadelphia}. Stimulants, like cocaine, and cannabis are commonly observed to co-occur with narcotics, and there is growing concern among the public and policymakers over an increasing number of unintentional overdose deaths involving stimulants. Marijuana and heroin are frequently mentioned together in discussions around legalization and concerns about adulteration from fentanyl. Our analysis also shows that narcotics and treatment drugs, including methadone, naloxone, and buprenorphine, are the second most commonly co-located drugs in articles. These articles primarily focus on criminality, legislature, and overdose, as determined by our TF-IDF analysis.
There is growing concern over the co-occurrence of stimulants, like cocaine, with narcotics, like heroin, due to the rising number of fatalities associated with their use \cite{johnson_increasing_2021}.
The Philadelphia Inquirer has been particularly active in discussing the ``opioid epidemic,'' with a focus on treatment. The use of treatment drugs like methadone and naloxone is seen as a positive and proactive response to drug addiction \cite{national_academies_of_sciences_effectiveness_2019}. Buprenorphine, another popular substance used to treat opioid use disorder, gained attention in 2022 after President Biden eliminated the restriction on the number of prescriptions providers could issue per month, making it more accessible compared to methadone \cite{buprenorphine_2021}.

\begin{figure}[h]
    \centerline{\includegraphics[scale=0.15]{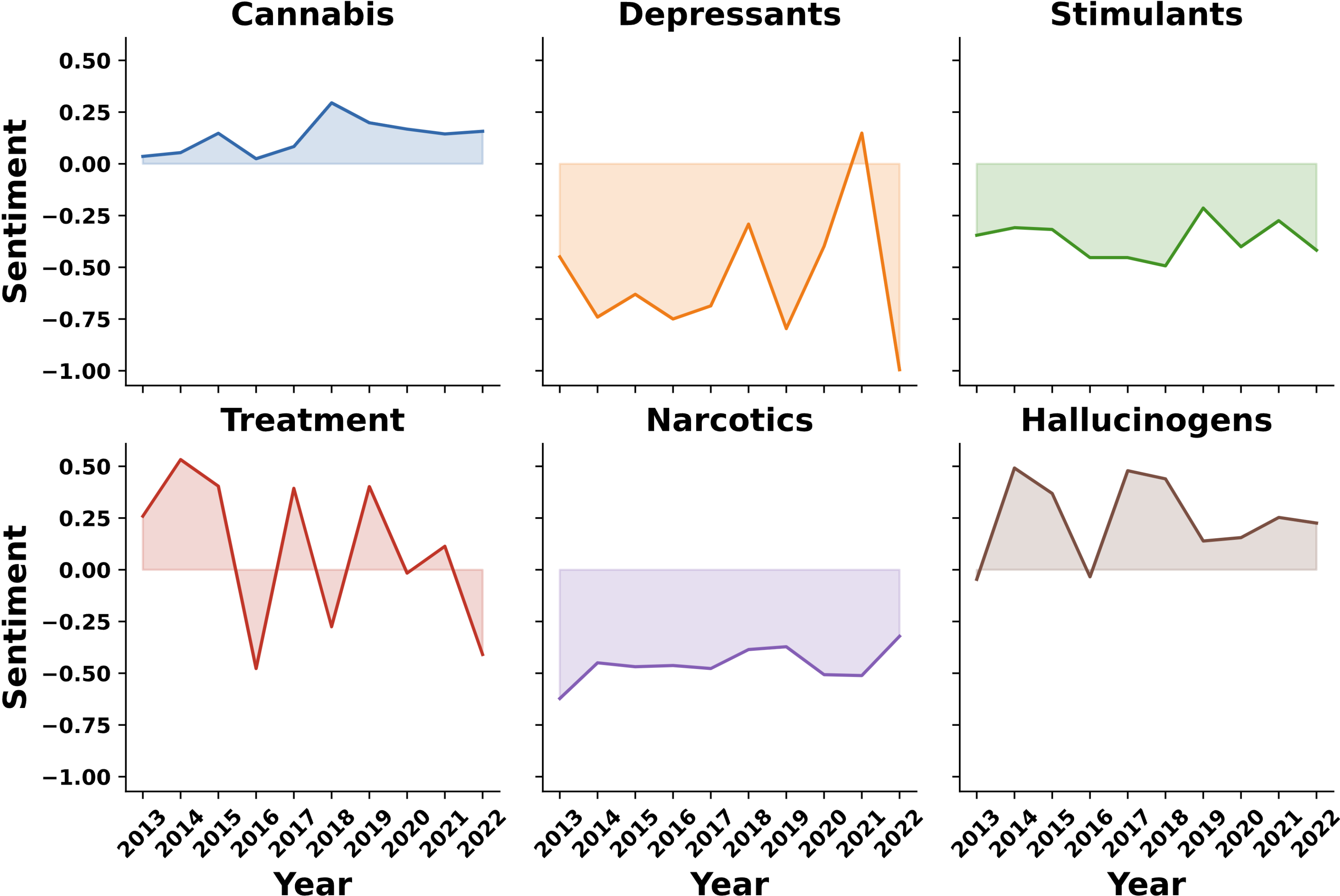}}
    \caption{Drug class sentiment scores over the past decade.}
    \label{fig:senti_time}
\end{figure}

\paragraph{How has media coverage of illicit substance use evolved over time?}
Utilizing dynamic topic modeling with BERTopic, we were able to generate a diverse set of topics covering various themes. The review of the initial results revealed a number of overlapping topics. Therefore, we iteratively refined the model by decreasing the number of topics until we achieved a high-quality output. Figure \ref{fig:topics_over_time} presents temporal changes of topic clusters in our data. The result shows that cannabis-related topics are the most prevalent in our dataset, with a specific focus on legislation. We also observe a significant number of news articles discussing drug use in films and television series, indicating the possible role of popular media in shaping public perceptions and attitudes toward drugs and social issues \cite{rezapour2017classification,diesner2016assessing}. 

The sentiment analysis of news articles demonstrates that the tone of news articles varies across drug classes and evolves over time (Figure \ref{fig:senti_time}).
News articles with hallucinogenic drugs tend to have a more positive tone compared to other classes. 
The positive tone may be attributed to recent studies on the use of psychedelic drugs for the treatment of mental health disorders such as treatment-resistant depression and post-traumatic stress disorder. For instance, we identified an article published in the Philadelphia Inquirer discussing the use of LSD and shrooms to treat PTSD \cite{ao2020can}. This article reflects the positive sentiment toward the therapeutic potential of hallucinogenic drugs. 
Furthermore, the majority of news articles mentioning cannabis, hallucinogens, and treatment drugs exhibit positive sentiments, whereas news articles about narcotics, stimulants, and depressants depict negative sentiments. The sentiment scores assigned to each article range from -1 (most negative) to +1 (most positive), and neutral sentiments are rarely observed in our dataset

Overall, our analysis indicates a shift in the discussion surrounding drug-related topics, especially hallucinogens, towards a more positive view of medicated-assisted treatment and novel applications of these substances. This emphasizes the importance of reporting drug-related incidents using harm reduction principles to encourage safer and managed use. 
Further research is necessary to fully understand the context and implications of these findings. It is also important to acknowledge that our analysis only captures a subset of the wider discussion surrounding these drugs and that additional studies may reveal further insights.

%% file: conclusion.tex
\section{Conclusion and Future Work}
Over the past decade, there has been an evolution in how the media portrays illicit substance use, as our study has revealed. We analyzed news articles published in the Philadelphia Inquirer between 2013 and 2022 and found that cannabis was the most frequently discussed drug class, followed by narcotics. News articles about hallucinogenic drugs tend to have a more positive tone compared to other categories of drugs, while articles on narcotics were the most negative. By examining changes in the tone and frequency of drug-related discussions over time, we can gain a better understanding of how societal attitudes towards drugs and drug policies are evolving, and how this may impact public health and well-being. It is important to note that our study focused solely on news articles published in the Philadelphia Inquirer, and therefore our findings are specific to this source and its coverage of substance use within the Philadelphia and United States. Further research is needed to explore how the portrayal of substance use in the media varies across different regions and cultures.